\documentclass[sigconf]{acmart}
\usepackage{multirow}

\usepackage{balance}

\AtBeginDocument{%
  }

\copyrightyear{2023}
\acmYear{2023}
\setcopyright{acmlicensed}
\acmConference[MM '23] {Proceedings of the 31st ACM International Conference on Multimedia}{October 29--November 3, 2023}{Ottawa, ON, Canada.}
\acmBooktitle{Proceedings of the 31st ACM International Conference on Multimedia (MM '23), October 29--November 3, 2023, Ottawa, ON, Canada}
\acmPrice{15.00}
\acmISBN{979-8-4007-0108-5/23/10}
\acmDOI{10.1145/3581783.3611724}

\begin{document}


\title{CATR: Combinatorial-Dependence Audio-Queried Transformer for Audio-Visual Video Segmentation}



\author{Kexin Li}
\affiliation{%
  \institution{Zhejiang University}
  \city{Hangzhou}
  \country{China}}
\email{12221004@zju.edu.cn}


\author{Zongxin Yang}
\orcid{0000-0001-8783-8313}
\authornote{Zongxin Yang is the corresponding author.}
\affiliation{%
  \institution{Zhejiang University}
  \city{Hangzhou}
  \country{China}}
\email{yangzongxin@zju.edu.cn}

\author{Lei Chen}
\affiliation{%
\institution{Finvolution Group}
\city{Shanghai}
\country{China}}
\email{chenlei04@xinye.com}

\author{Yi Yang}
\affiliation{%
\institution{Zhejiang University}
\city{Hangzhou}
\country{China}}
\email{yangyics@zju.edu.cn}

\author{Jun Xiao}
\affiliation{%
\institution{Zhejiang University}
\city{Hangzhou}
\country{China}}
\email{junx@cs.zju.edu.cn}
\renewcommand{\shortauthors}{Kexin Li, Zongxin Yang, Lei Chen, Yi Yang, \& Jun Xiao}

\begin{abstract}

  Audio-visual video segmentation~(AVVS) aims to generate pixel-level maps of sound-producing objects within image frames and ensure the maps faithfully adhere to the given audio, such as identifying and segmenting a singing person in a video. However, existing methods exhibit two limitations: 1) they address video temporal features and audio-visual interactive features separately, disregarding the inherent spatial-temporal dependence of combined audio and video, and 2) they inadequately introduce audio constraints and object-level information during the decoding stage, resulting in segmentation outcomes that fail to comply with audio directives. To tackle these issues, we propose a decoupled audio-video transformer that combines audio and video features from their respective temporal and spatial dimensions, capturing their combined dependence. To optimize memory consumption, we design a block, which, when stacked, enables capturing audio-visual fine-grained combinatorial-dependence in a memory-efficient manner. Additionally, we introduce audio-constrained queries during the decoding phase. These queries contain rich object-level information, ensuring the decoded mask adheres to the sounds. Experimental results confirm our approach's effectiveness, with our framework achieving a new SOTA performance on all three datasets using two backbones. The code is available at \url{https://github.com/aspirinone/CATR.github.io}.

\end{abstract}


\maketitle

\begin{figure*}[t]
    \centering
    \includegraphics[width=1\textwidth]{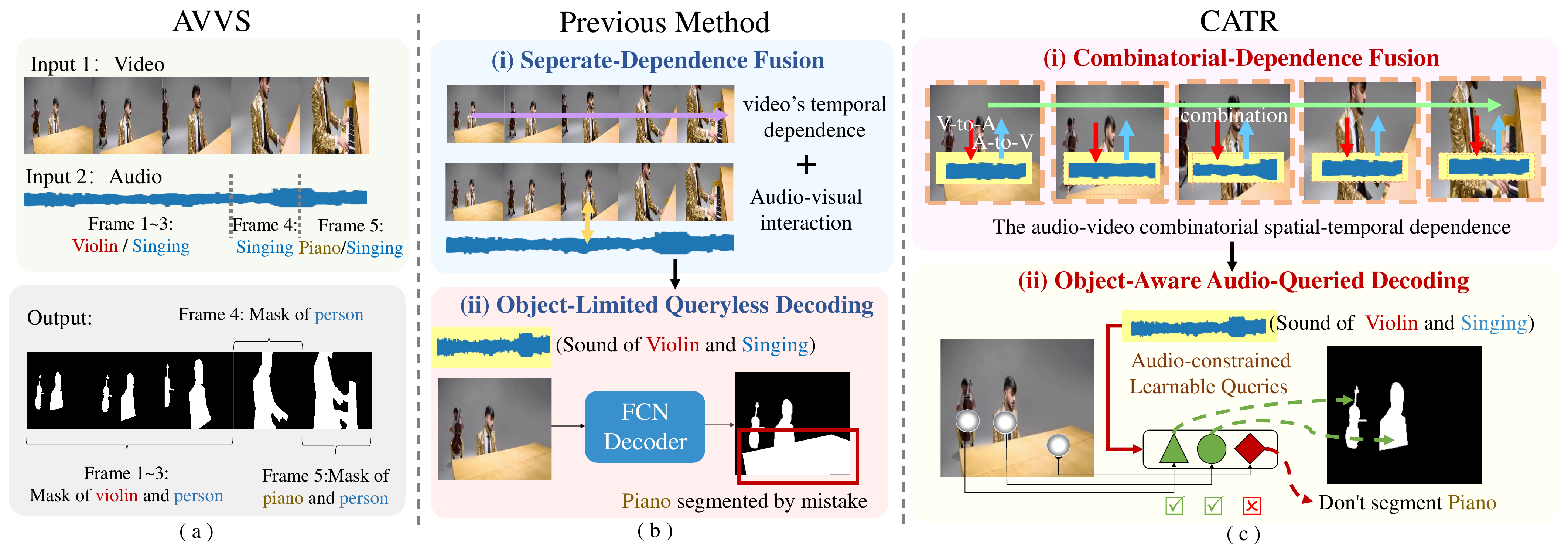}
    \caption{{\bf AVVS Task Description and CATR Contributions. The objective of the Audio-Visual Video Segmentation (AVVS) task is to generate pixel-level maps identifying sound-producing objects within image frames~(a). Previous approaches separately addressed the temporal dependencies of video and the audio-video interaction information~(b), neglecting the unique spatial-temporal dependencies inherent to audio and video as a combination. CATR initially merges audio and video features, subsequently capturing the spatial-temporal dependencies of this combination. Note that the red arrow symbolizes the Video-to-Audio~(V-to-A) information, while the blue arrow denotes the Audio-to-Video~(A-to-V) information. Additionally, we introduce innovative audio-constrained learnable queries to enhance object-aware segmentation~(c).} }
    \label{fig:introduction} 
\end{figure*}

\section{Introduction}
Audio-visual video segmentation~(AVVS) aims to generate pixel-level maps of sound-producing objects within image frames and ensure the maps faithfully adhere to the given audio. For instance, when a person sings, AVVS enables the identification and segmentation of individuals in the video~(see Figure~\ref{fig:introduction}~(a)). This capability has significant implications for various applications, such as video editing and surveillance. Despite the successful integration of multi-modal guidance approaches using point, box, scribble, text, and verbal cues for segmentation in recent studies~\cite{samtrack, kirillov2023segment, zou2023segment}, audio guidance has not yet been incorporated. This gap can be attributed to the inherent challenges of AVVS, including the ambiguous semantic information embedded in sounds and establishing correspondence between sounds and pixel-level predictions. Therefore, future research needs to investigate the integration of multi-knowledge representations~\cite{mkr}, including audio, video, segmentation, etc.

In the domain of referring video object segmentation~\cite{wu2022language,zou2023segment,liang2023local} and audio-visual understanding~\cite{lin2019dual, lin2020audiovisual, tian2018audio, wu2019dual, xu2020cross, xuan2020cross, lin2019dual, ramaswamy2020see, duan2021audio, zhou2021positive,tian2020unified, wu2021exploring, lin2021exploring, yu2022mm}, substantial efforts have been devoted to investigating multi-modal segmentation techniques. Zhou et al.~\cite{DBLP:conf/eccv/ZhouWZSZBGKWZ22}, for example, introduced a framework incorporating the TPAVI module, which facilitates audio-visual pixel-level segmentation. However, these methods still encounter two primary limitations in the realms of audio fusion and audio-guided video decoding:

Firstly, \textbf{Seperate-Dependence Fusion.} The challenges in utilizing audio features arise from the ambiguous semantic information embedded within sounds, such as differentiating a child's cry from a cat's meow, in contrast to the clear linguistic references to "child" and "cat". As a result, establishing precise pixel-level associations under these indeterminate auditory cues is difficult. However, we found that audio has a unique advantage: its temporal properties align with those of video features, capturing distinct but complementary aspects of the same event. Existing methods do not fully exploit this property, addressing video temporal information and audio-visual interactions separately, which constrains their effectiveness. Various combinations of audio and video exhibit unique spatial-temporal dependencies, contributing to more accurate and robust results. Thus, a method that captures the spatial-temporal characteristics of audio and video in combination is essential.

Secondly, \textbf{Object-Limited Queryless Decoding.} Previous methods typically derive the final mask directly after decoding video features, as exemplified by the use of an FCN decoder~\cite{DBLP:conf/eccv/ZhouWZSZBGKWZ22}. This approach neglects audio guidance information and omits object-level information during the decoding stage, potentially leading to segmentation errors in complex environments. For instance, in Figure~\ref{fig:introduction}~(b), the second frame of the video contains a violin, a piano, and people simultaneously. With audio containing only violin sounds and human singing, the target segmentation objects should be the person and the violin. However, due to the absence of audio constraints during the decoding phase, previous methods may erroneously segment the piano, influenced by the video's focus on the front and back frames. Consequently, it is essential to introduce audio restrictions and provide object-level guidance information during the decoding phase.

To address the above limitations, we designed targeted modules:

(1) \textbf{Combinatorial-Dependence Fusion.} To comprehensively assess audio-visual combinatorial-dependence, we design to combine audio and video features from their respective temporal and spatial dimensions, followed by capturing this combination's spatial-temporal dependence. Commonly, transformers are used to capture temporal dependencies; assuming a video frame dimension of $H \times W$, the merged feature dimension becomes $(H \times W + T)$. However, due to the substantial memory consumption associated with this encoder, we propose an innovative decoupling transformer that considerably reduces memory usage while allowing the extraction of spatial-temporal interaction information between audio-audio, video-video, audio-video, and video-audio combinations. 

(2) \textbf{Object-Aware Audio-Queried Decoding.} To enable attention focus on the object of interest, we propose an audio-queried decoder. Specifically, we apply an audio constraint to all object queries, allowing the model to leverage audio information to direct attention toward the desired object. These conditional queries serve as inputs for the model, which produces object-aware dynamic kernels to filter segmentation masks from feature maps.

\begin{figure*}[t]
    \centering
    \includegraphics[width=1\textwidth]{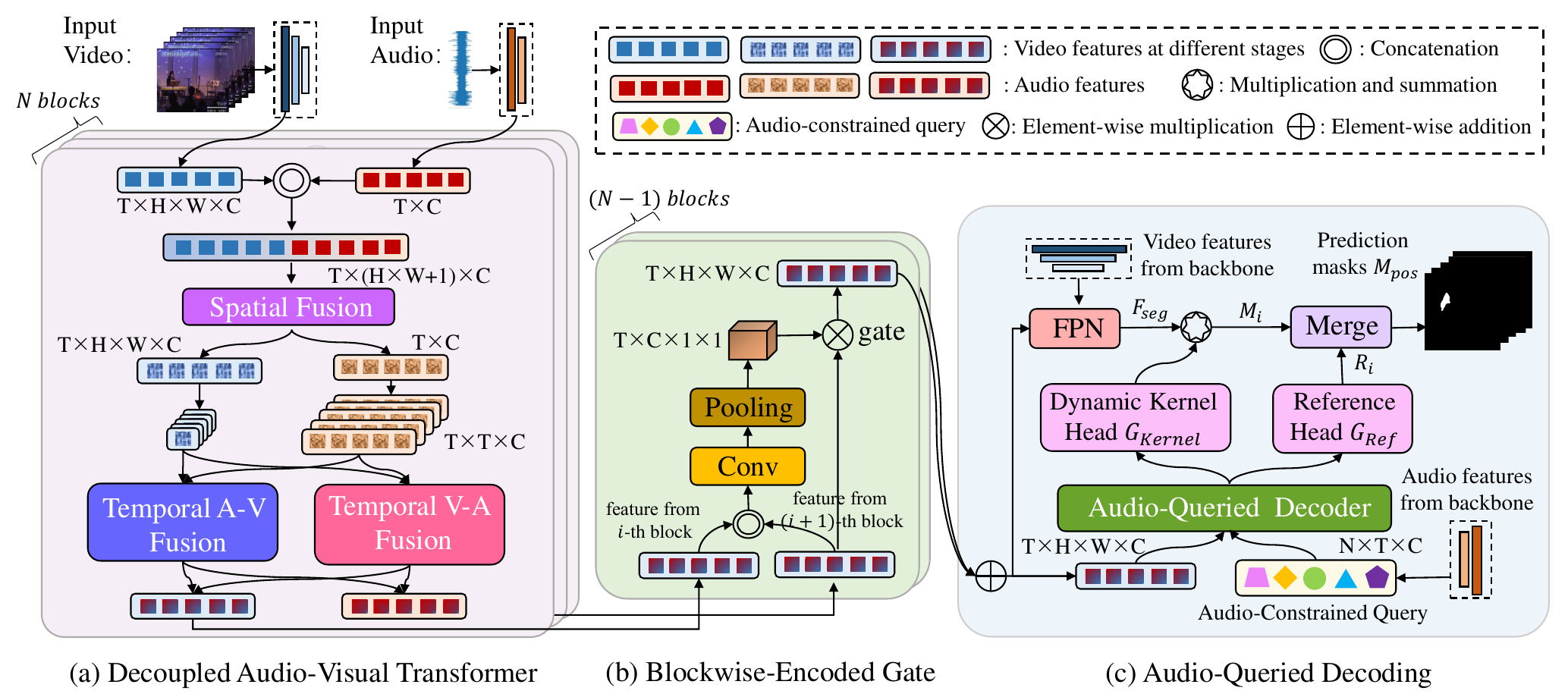}
    \caption{{\bf CATR employs an encoder-decoder structure. (a) In encoding, we merge audio and video features and capture their spatial-temporal combinatorial-dependencies. To conserve memory, we devise decoupling methods, utilizing temporal A-V and temporal V-A to fusion audio and video features. (b) To balance the contributions of multiple encoder blocks, we implement a blockwise gating method for utilizing all video features from each block. (c) In decoding, we introduce audio-constrained learnable queries, which harness audio features to extract object-level information, guiding target object segmentation.} }
    \label{fig:pipeline} 
\end{figure*}

On the whole, we propose a \textbf{C}ombinatorial-Dependence \textbf{A}udio-Queried \textbf{T}ransforme\textbf{r} Network~(\textbf{CATR}; Figure.~\ref{fig:pipeline}), which contains two main components: Decoupled Audio-Visual Transformer Encoding Module~(DAVT; detailed in Section.~\ref{sec:fusion}) and Audio-Queired Decoding Module~(detailed in Section.~\ref{sec:decoder}). In encoding, we design an innovative decoupling block, which consists of two steps: initially, we merge audio and video features of corresponding frames while concurrently capturing their temporal information. Subsequently, we facilitate interaction between video features containing temporal information and audio features. By stacking decoupling blocks, we can efficiently capture audio-visual spatial-temporal correlations in a memory-efficient manner. In addition, Audio-Queried Decoder Module innovatively employ an audio constraint to all object queries to produce object-aware dynamic kernels to filter the segmentation of desired object. Moreover, we design a Blockwise-Encoded Gate to utilize all the features extracted from each encoder block. This Blockwise-Encoded Gate enables modeling of the overall distribution of all encoder blocks from a global perspective, thereby balancing the contributions of different encoder blocks.

We conduct extensive experiments on three popular benchmarks and achieve new state-of-the-art performance on all datasets with two backbones~(On S4, CATR 84.4\% $\mathcal {J}$ / 91.3\% $\mathcal {F}$ vs. TPAVI 78.7\% $\mathcal {J}$ / 87.9\% $\mathcal {F}$; On M3, CATR 61.8\% $\mathcal {J}$ / 71\% $\mathcal {F}$ vs. TPAVI 54\% $\mathcal {J}$ / 64.5\% $\mathcal {F}$). Our code and benchmark will be released.

Overall, our contributions are summarized as follows:
\begin{itemize}
\vspace{-0.1cm}
\setlength{\itemsep}{0pt}
\setlength{\parsep}{0pt}
\setlength{\parskip}{0pt}
\item We introduce an encoding-decoding framework CATR that presents a novel spatial-temporal audio-video fusion block to fully consider the audio-visual combinatorial dependence in a decoupled and memory-efficient manner.

\item We propose the audio-constrained learnable queries to incorporate audio information comprehensively during decoding. These audio-constrained queries contain abundant object-level information that can select which object is being referred to segment. In addition, we introduce a Blockwise-Encoded Gate that allows for the selective fusion of features from different encoder blocks.

\item We conduct extensive experiments on three popular benchmarks, and achieve new superior state-of-the-art performance on all three datasets with two backbones.
\end{itemize}

\section{Related Work}
\subsection{Video Object Segmentation (VOS)}
The VOS task~\cite{wang2021survey,cfbip} aims to segment the object of interest throughout the entire video sequence. It is divided into two settings: semi-supervised and unsupervised. For semi-supervised VOS~\cite{yang2021associating,yang2022decoupling,zhang2023boosting,lu2020video}, the target object is decided given a one-shot mask of the first video frame. As for unsupervised VOS~\cite{li2023unified}, it needs to segment all the primary objects automatically. Many excellent works are proposed and proven to achieve impressive segmentation performance. However, these fancy designs are limited to a single visual modality. 

\subsection{Audio-Visual Video Segmentation (AVVS)}
The human ability to identify objects is not solely reliant on visual cues but also on auditory signals. For instance, the distinct sounds of a dog barking or a bird chirping are easily recognizable. This observation underscores the complementarity of audio and visual information. However, while speech-guided video segmentation is a more reliable means of distinguishing instance-level objects, sound can only provide information about object categories, making it a challenging task to locate and segment the object producing the sound. Zhou et al.~\cite{DBLP:conf/eccv/ZhouWZSZBGKWZ22} pioneered the audio-visual segmentation (AVVS) task and proposed a framework incorporating the TPAVI module, a groundbreaking approach for achieving pixel-level segmentation using audio information. Nonetheless, their framework's handling of multi-modal feature fusion and audio guidance was inadequate. Thus, we present a novel framework that addresses these limitations.

\subsection{Vision Transformers}
Transformer~\cite{DBLP:conf/nips/VaswaniSPUJGKP17} was first introduced for sequence-to-sequence translation in natural language processing community and has achieved marvelous success in most computer vision tasks~\cite{DBLP:conf/iclr/DosovitskiyB0WZ21, DBLP:conf/iccv/KamathSLSMC21, DBLP:conf/iccv/LiuL00W0LG21, li2023commonsense} such as object detection~\cite{DBLP:conf/eccv/CarionMSUKZ20, DBLP:conf/iclr/ZhuSLLWD21}, tracking~\cite{DBLP:conf/cvpr/ChenYZ0YL21, DBLP:conf/cvpr/MeinhardtKLF22, DBLP:journals/corr/abs-2012-15460, DBLP:conf/iccv/0002PF0L21} and segmentation~\cite{yang2022decoupling,DBLP:conf/nips/ChengSK21, DBLP:conf/cvpr/ZhengLZZLWFFXT021,liang2022gmmseg}. The Transformer employs an attention mechanism to facilitate the transformation of input into output representations. Building upon this foundation, the DETR~\cite{DBLP:conf/eccv/CarionMSUKZ20} has advanced the field by introducing a learnable query mechanism, which serves to expand the range of output possibilities. By employing an intelligent query and output matching mechanism, DETR is capable of determining the most optimal association between input and output elements. Furthermore, the VisTR~\cite{DBLP:conf/cvpr/WangXWSCSX21} extends the capabilities of DETR to the domain of video segmentation, achieving notable advancements. DeAOT~\cite{yang2022decoupling} decouples the visual and identification features in hierarchical propagation~\cite{yang2021associating} and achieves state-of-the-art performance in semi-supervised VOS. Inspired by these works, our work also relies on the query mechanism of Transformer but considers an additional modality, i.e., audio, as the object reference. Moreover, we propose an effective spatial-temporal fusion module to realize audio-guided video segmentation.

\section{Method}
\subsection{Overview}
Our pipeline for AVVS task can be formulated as encoding-decoding (depicted in Figure.~\ref{fig:pipeline}). To address limitations in previous methods, such as inadequate correlation and vague reference, we carefully design two modules: the Decoupled Audio-Visual Transformer Encoding Module~(DAVT; detailed in Section.\ref{sec:fusion}) and the Audio-Queired Decoding Module~(detailed in Section.\ref{sec:decoder}). These modules enable effective audio-visual spatial-temporal connection and capture the object-level information to achieve more explicit reference, respectively. Moreover, we design a Blockwise-Encoded Gate to enable modeling of the overall distribution of all encoder blocks. In addition, CATR aims to output a pixel-level map of the object(s) that produce sound at the time of the image frame,
\begin{equation}
\{M_{t}\}_{t=1}^T = CATR({S_{t,v}}, {S_{t,a}}),
\end{equation}
where we denote the video sequence as $S=\{S_{t,v}, S_{t,a}\}_{t=1}^T$. Moreover, $S_v$ denotes the visual sequence and $S_a$ denotes the audio sequence. The predictions are denoted as $\{M_t\}_{t=1}^T$,  ${M_t} \in \mathbb{R}^{H \times W}$.

\section{Decoupled Audio-Visual Transformer}
\label{sec:fusion}
In contrast to existing methods that account for the video temporal information and audio-visual interaction separately, we propose a method that obtains the spatial-temporal combinatorial-dependence between audio-audio, video-video, audio-video, and video-audio in a novel decoupling memory-efficient manner. 
~\\
 
\noindent {\bf{Stack DAVT Blocks.}}
To conserve memory, we designed the Decoupled Audio-Visual Transformer ~(DAVT). The DAVT block involves two steps. Initially, we combine the audio and video features of corresponding frames and capture their temporal information simultaneously. Subsequently, we interact processed video features with audio features, respectively.

For a video sequence $S_v$, we extract visual features after popular backbones and atrous spatial pyramid pooling~\cite{chen2017deeplab} and obtain hierarchical visual feature maps. We denote the video features as ${F_v}\in \mathbb{R}^{T \times H \times W \times C}$, where $T$, $H$, $W$ and $C$ signifying the number of frames, height, width, and channel, respectively. Given an audio sequence $S_a$, we employ a convolutional neural network VGGish ~\cite{DBLP:conf/icassp/HersheyCEGJMPPS17} pre-trained on AudioSet~\cite{gemmeke2017audio} as backbone to extract audio features ${F_a} \in \mathbb{R}^{T \times d}$, where $d = 128$. 
\begin{equation}
{{F}_v^{l+1}, {F}_a^{l+1}} = DAVT({F}_v^{l}, {F}_a^{l}),
\end{equation}
where $DAVT(\cdot)$ denotes the Decoupled Audio-Visual Transformer block, $l$ denotes the $l$-th block. By stacking multiple DAVT blocks, we can effectively capture the spatial-temporal correlation between audio-audio, video-video, audio-video and video-audio in a memory-efficient manner.
~\\

\noindent {\bf{Spatial Audio-Visual Fusion.}}
To obtain the audio-visual overall dependence, the visual features $F_v^l$ and audio features $F_a^l$ are linearly projected to a shared dimension $D$. The video features for each frame are flattened and individually merged with the audio embeddings, yielding a set of $T$ multi-modal sequences, each of shape $(H \times W + 1) \times D$. 
\begin{equation}
{\tilde{{F}}_v^{l}, \tilde{{F}}_a^{l}} = SF(Concat({F}_v^{l}, {F}_a^{l})),
\end{equation}
where $Concat(\cdot)$ denotes the concatenate operation, and $SF(\cdot)$ denotes the spatial audio-visual fusion function, which is employed as self-attention. Then we obtain the processed video feature $\tilde{{F}}_v^{l}$ that contains the corresponding frame audio information. Similarly, the audio feature $\tilde{{F}}_a^{l}$ contains the corresponding frame video information.
~\\

\noindent {\bf{Temporal A-to-V Fusion.}}
Employing a transformer-based encoder will consume a large amount of memory, so we use the decoupling method to carry out Audio-to-Video~(A-to-V) interaction and Video-to-Audio~(V-to-A) interaction respectively. 
\begin{equation}
{\hat{{F}}_v^{l}, \hat{{F}}_a^{l}} = TAV(\tilde{{F}}_v^{l}, \tilde{{F}}_a^{l}) = \operatorname{Softmax}\left(\frac{\tilde{{F}}_v^{l} W^Q \cdot\left(\tilde{{F}}_a^{l} W^K\right)^{\mathrm{T}}}{\sqrt{d_{\text {head }}}}\right) \tilde{{F}}_a^{l} W^V
\end{equation}
where the $TAV(\cdot)$ denotes the Temporal Audio-to-Video Fusion, which is employed as multi-head attention~\cite{vaswani2017attention}. In $TAV(\cdot)$, the query is the processed video feature $\tilde{{F}}_v^{l}$, and key is the audio feature $\tilde{{F}}_a^{l}$. Moreover, $W^Q, W^K, W^V \in \mathbb{R}^{C \times d_{\text {head }}}$ are learnable parameters. 
~\\

\noindent {\bf{Temporal V-to-A Fusion.}}
Correspondingly, we also design a Temporal Video-to-Audio Fusion function $TVA(\cdot)$,
\begin{equation}
{\check{{F}}_v^{l}, \check{{F}}_a^{l}} = TVA(\tilde{{F}}_v^{l}, \tilde{{F}}_a^{l}) = \operatorname{Softmax}\left(\frac{\tilde{{F}}_a^{l} W^Q \cdot\left(\tilde{{F}}_v^{l} W^K\right)^{\mathrm{T}}}{\sqrt{d_{\text {head }}}}\right) \tilde{{F}}_v^{l} W^V,
\end{equation}
where $TVA(\cdot)$ denotes the Temporal Video-to-Audio Fusion that is also employed as multi-head attention. In $TVA(\cdot)$, the query is the processed audio feature $\tilde{{F}}_a^{l}$ and key is the video feature $\tilde{{F}}_v^{l}$. 

After we obtain the video feature $\check{{F}}_v^{l}$ that from $TVA(\cdot)$ and $\hat{{F}}_v^{l}$ that from $TAV(\cdot)$, we merge the $\check{{F}}_v^{l}$ and $\hat{{F}}_v^{l}$ by element-wise adding and obtain the fully interacted video feature $\bar{{F}}_v^{l}$
~\\

\begin{figure}[t]
    \centering
    \includegraphics[width=0.45\textwidth]{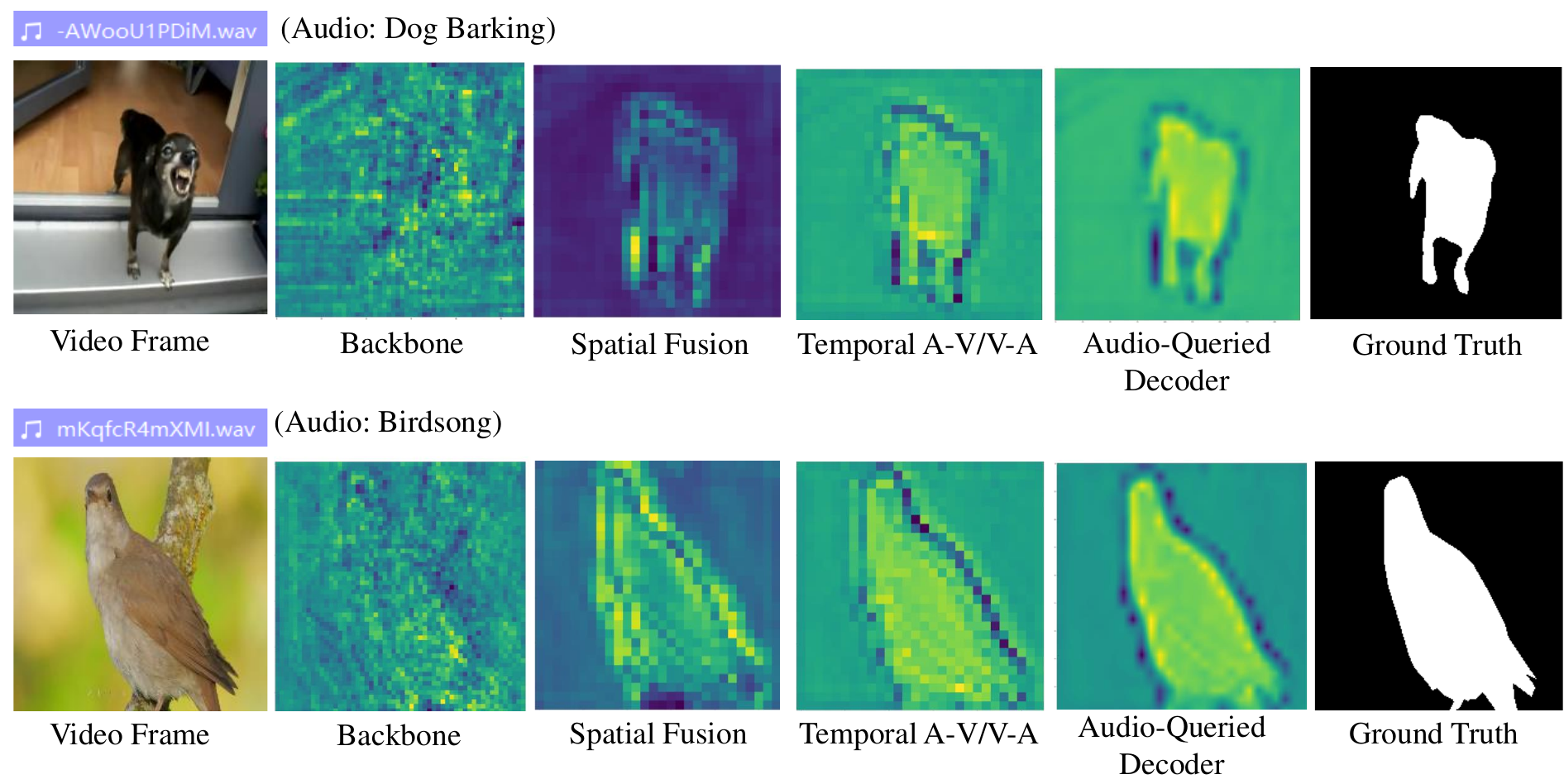}
    \vspace{-0.3cm}
    \caption{Visualization of video features after processing at each stage. We observed that the initial features, generated by the backbone network, appeared indistinct. However, the video features progressively aligned with the desired segmentation object after the spatial-temporal encoding module.}
    \label{fig:feature} 
    \vspace{-0.3cm}
\end{figure}

\noindent {\bf{Blockwise-Encoded Gate.}}
The existing method typically employs the features of the last encoder block alone as the decoder input, which is insufficient because the features of each encoder block contain varying degrees of multi-modal interaction information~(see Figure~\ref{fig:feature}). Thus, we design gate mechanisms to utilize all the features extracted from each encoder block and balance the contributions of different encoder blocks.

\begin{figure}[t]
    \centering
    \includegraphics[width=0.45\textwidth]{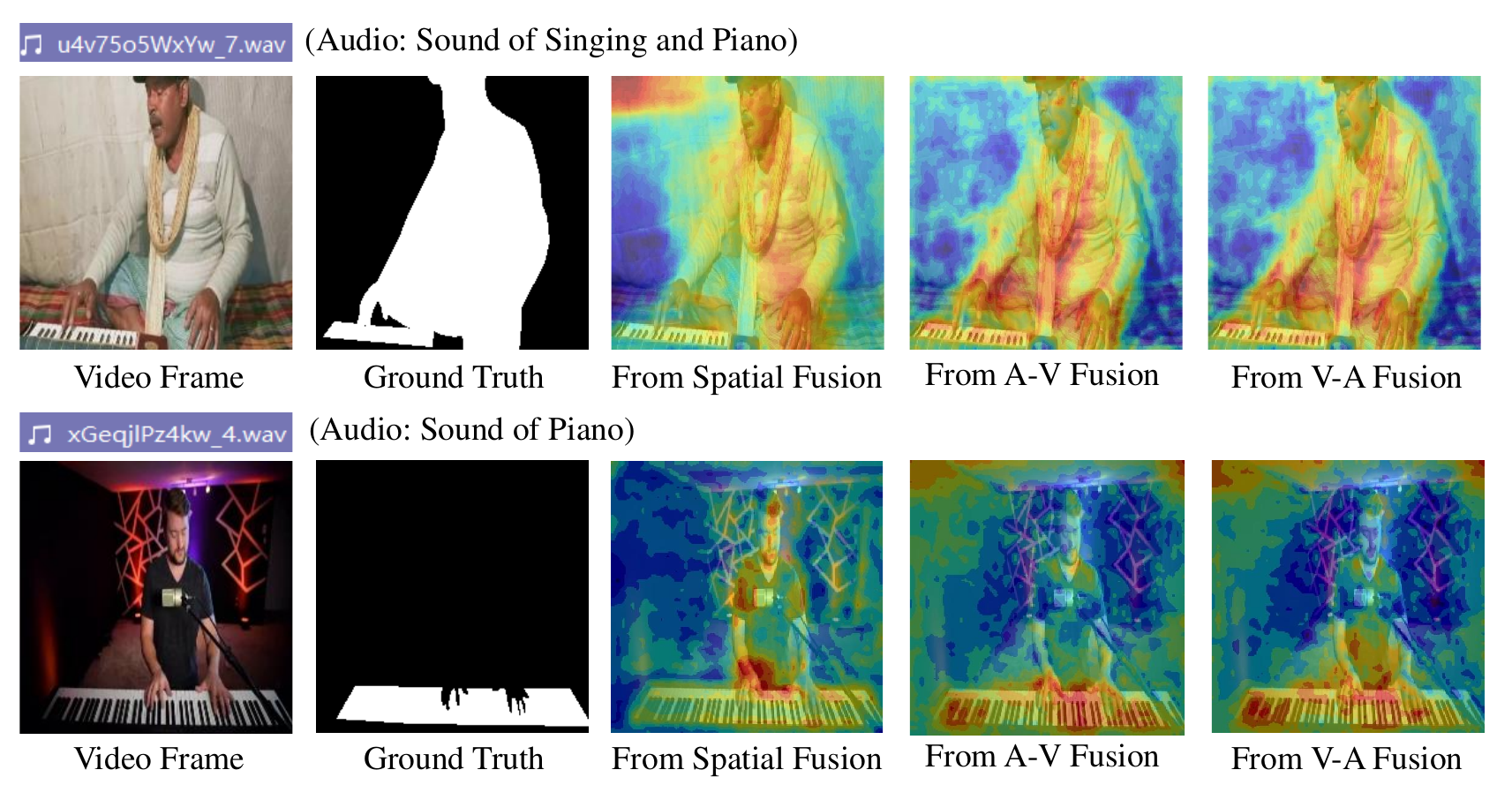}
    \vspace{-0.3cm}
    \caption{{\bf Attention maps generated from spatial fusion \& temporal A-V/V-A fusions. Sample 1: target is person \& piano; spatial fusion focuses on person, neglects piano; with temporal A-V/V-A fusions, the attention map accurately highlights both. Sample 2: target is the piano; spatial fusion wrongly emphasizes person, but temporal A-V/V-A maps correctly focus on piano. Consequently, we draw the conclusion that spatial fusion provides an initial integration of audio information, whereas temporal A-V and V-A fusions further consolidate this information to accurately identify the target object.} }
    \label{fig:attention} 
    \vspace{-0.3cm}
\end{figure}

Suppose we have two video features $\bar{{F}}_v^{l}$ and $\bar{{F}}_v^{l+1}$ from different Spatial-Temporal Encoding blocks, we design a gate unit and ${G}^{l+1}$ denotes the $(l+1)$-th output vector,
\begin{equation}
\begin{aligned}
\begin{split}
{G}^{l+1} &= Pool(Sigmoid(Conv(Concat(\bar{{F}}_v^{l},\bar{{F}}_v^{l+1})))), \\
{{F}^{final}_v} &= Conv({G^l} \cdot \bar{{F}}_v^{l} + {G^{l+1}} \cdot \bar{{F}}_v^{l+1}),
\end{split}
\end{aligned}
\end{equation}
where $Concat(\cdot)$ denotes the concatenate operation, $Conv(\cdot)$ denotes the convolution layer, $Sigmoid(\cdot)$ denotes the sigmoid function and $Pool(\cdot)$ denotes the global average pooling. The output channel of $Conv(\cdot)$ is $C$, which means the resulted gate vector ${G}^{l+1}$ has $C$ different elements which correspond to $C$ gate values~(we set $C=256$ here).

The gate values ${G}^{l+1}$ is applied for weighting the different-blocks video features $\bar{{F}}_v^{l}$ and $\bar{{F}}_v^{l+1}$. To obtain the final video encoding feature ${F}^{final}_v$, we fuse all the re-weighted features by element-wise addition and convolutional layers.

\subsection{Audio-Queired Decoding}

The existing methods fall short in effectively capturing object-level details and offering explicit information for cross-modal reasoning. To overcome this limitation, we propose audio-constrained queries, which impose an audio constraint on all object queries and generate object-aware dynamic kernels that filter target object segmentation masks from feature maps. Our approach aims to provide a comprehensive solution that enhances object recognition by incorporating audio signals into the process.
~\\

\label{sec:decoder}
\noindent {\bf{Audio-Constrained Query.}}
We hierarchically fuse the final video feature ${F}^{final}_v$ and the multi-layer features from backbone with an FPN-like~\cite{DBLP:conf/cvpr/LinDGHHB17} decoder, then we obtain the semantically-rich video feature maps ${F_{seg}} = \{ f_{seg,t} \}_{t=1}^T$, where $f_{seg,t} \in \mathbb{R}^{\frac{H}{8} \times \frac{W}{8} \times C}$.

To capture the object-level information comprehensively, we devised a set of $N$ learnable queries. These queries, along with the audio feature, were fed into the decoder embedding and position embedding in the transformer, resulting in queries with abundant object-level information. Next, we use two-layer dynamic kernels $\mathcal{G}_\mathrm{kernel}$ to generate a corresponding sequence segmentation for each query. Finally, the binary masks are generated by dynamic convolution:
\begin{equation}
{M_{i}} = \{ {F_{seg}}* \omega_i \}_{i=1}^{N},
\end{equation}
where $M_{i} \in \mathbb{R}^{N \times \frac{H}{8} \times \frac{W}{8}}$ denotes the segmentation mask with $N$ queries. $\omega_i$ and $F_{seg}$ denote the $i$-th dynamic kernel weights and its exclusive feature map, respectively.
~\\

\noindent {\bf{Query Matching.}}
The aim of query matching is to determine which of the predicted sequences best fits the referred object. Here, we denote each ground-truth sequence as $y=(M, R)=(\{M_{t}\}_{t=1}^{T},\{R_{t}\}_{t=1}^{T})$, where $M$ denotes the ground-truth mask and $R$ denotes a probability scalar indicating whether the instance corresponds to the referenced object and ascertains the visibility of this object within the current frame. In addition, we denote the prediction set as $\hat{y}= \{\hat{y}_i\}_{i=1}^N$,where $ \hat{y}_i =  (\{\hat{M}_{i,t}\}_{t=1}^{T},\{\hat{R}_{i,t}\}_{t=1}^{T})$.

To find the best prediction from $N$ conditional queries, we use a reference head $\mathcal{G}_\mathrm{Ref}$, which consists of a single linear layer followed by a softmax layer. Then we obtain the positive sample by minimizing the matching cost:
\begin{equation}
\begin{aligned}
\begin{split}
\hat{y}_{\mathrm{pos}}&=\underset{\hat{y}_i \in \hat{y}}{\arg \min } \;
 \mathcal{C}_{\text {match }}\left(y, \hat{y}_i\right), \\
 \mathcal{C}_{\text {match }}\left(y, \hat{y}_i\right) &= \mathcal{C}_{\text {dice }}(M, \hat{M}_i) + \mathcal{C}_{\text {ref }}(R, \hat{R}_i)
 \end{split}
\end{aligned}
\end{equation}
where $\hat{y}_{\mathrm{pos}}$ denotes the permutation in $N$ conditional queries with the lowest total cost. $\mathcal{C}_{\text {dice }}$ takes on the role of overseeing and evaluating the predicted mask sequence in direct comparison with the ground-truth mask sequence, with this evaluation process being conducted by the Dice coefficients~\cite{milletari2016v}, and $\mathcal{C}_{\text {ref }}$ utilizes cross-entropy to guide the reference predictions, aligning them with the corresponding ground-truth reference identity. 

\subsection{Loss and Inference}
We consider both mask and reference identity, and we define our loss function as follows:
\begin{equation}
\begin{aligned}
\begin{split}
&\mathcal{L}(y,\hat{y}_i)=\mathcal{L}_{\mathrm{Mask}}\left( M_i, \hat{M}_i\right)+\mathcal{L}_{\mathrm{Ref}}\left(R_i, \hat{R}_i\right) \\
&= \lambda_d \mathcal{L}_{\text {Dice }}(M_i, \hat{M}_i )+\lambda_f \mathcal{L}_{\text {Focal }}(M_i, \hat{M}_i) + \lambda_r \mathcal{L}_{\mathrm{Ref}}(R_i, \hat{R}_i)
\end{split}
\end{aligned}
\end{equation}
where $\mathcal{{L}}_{\mathrm{Mask}}$ ensures mask alignment between the predicted and ground-truth, and $\mathcal{L}_{\mathrm{Ref}}$ supervises the reference identity predictions. In addition, $\mathcal{L}_{\mathrm{Mask}}$ is implemented by a combination of the Dice~\cite{milletari2016v} and the per-pixel Focal~\cite{lin2017focal} loss functions, and $\mathcal{L}_{\mathrm{Ref}}$ is implemented by a cross-entropy term.

For inference, CATR will predict $N$ object sequences. For each sequence, we obtain the predicted reference probabilities and the reference score set $P = \{p_i\}_{i=1}^N$,  We select the object sequence with the highest score and its index is denoted as $R_{pos}$, 
\begin{equation}
R_{pos} = \underset{i \in \{1,2,…,N\}}{\arg \max }\;p_i
\end{equation}
Finally, we return the final mask $M_{pos}$ that corresponds to $R_{pos}$.

\begin{table}[t]
\begin{tabular}{ccccccc}
\toprule
                       &                          &                            & \multicolumn{2}{c}{S4}                                    & \multicolumn{2}{c}{M3}                                    \\
\multirow{-2}{*}{Task} & \multirow{-2}{*}{Method} & \multirow{-2}{*}{Backbone} & $\mathcal {M_J}$                           & $\mathcal {M_F}$                           & $\mathcal {M_J}$                           & $\mathcal {M_F}$                           \\ \midrule
                       & LVS~\cite{chen2021localizing}                      & resnet18                   & 37.9                        & 51                          & 29.5                        & 33                          \\
\multirow{-2}{*}{SSL}  & MSSL~\cite{qian2020multiple}                     & resnet18                   & 44.9                        & 66.3                        & 26.1                        & 36.3                        \\ \midrule
                       & 3DC~\cite{mahadevan2020making}                       & resnet152                  & 57.1                        & 75.9                        & 36.9                        & 50.3                        \\
\multirow{-2}{*}{VOS}  & SST~\cite{duke2021sstvos}                      & resnet101                  & 66.3                        & 80.1                        & 42.6                        & 57.2                        \\ \midrule
                       & iGAN~\cite{mao2021transformer}                      & resnet50                   & 61.6                        & 77.8                        & 42.9                        & 54.4                        \\
\multirow{-2}{*}{SOD}  & LGVT~\cite{zhang2021learning}                     & swin                       & 74.9                        & 87.3                        & 40.7                        & 59.3                        \\ \midrule
                       &                          & resnet50                   & 72.8                        & 84.8                        & 47.9                        & 57.8                        \\
                       & \multirow{-2}{*}{TPAVI~\cite{DBLP:conf/eccv/ZhouWZSZBGKWZ22}}  & PVT-v2                        & 78.7                        & 87.9                        & 54.0                          & 64.5                        \\
                       &                          & resnet50                   & 74.8                        & 86.6                        & 52.8                        & 65.3                        \\
                       & \multirow{-2}{*}{CATR}  & PVT-v2                        & 81.4                        & 89.6                        & 59.0                          & 70.0                          \\
                       &                          & resnet50                   & {\color{blue} 74.9} & {\color{blue} 87.1} & {\color{blue} 53.1} & {\color{blue} 65.6} \\
\multirow{-6}{*}{AVVS} & \multirow{-2}{*}{CATR*} & PVT-v2                       & {\color{red} 84.4}   & {\color{red} 91.3} & {\color{red} 62.7} & {\color{red} 74.5} \\ \bottomrule
\end{tabular}
\caption{Quantitative comparisons on AVSBench-object datasets~(single-source,S4; multi-source,M3). {\color{blue}{Blue}} indicates the best performance with resnet backbone, while {\color{red}{red}} indicates the best performance among all settings. * denotes that the training datasets are supplemented annotation with AOT.}
\label{tab:all}
\vspace{-0.3cm}
\end{table}

\begin{table}[t]
\begin{tabular}{ccccc}
\toprule
\multirow{2}{*}{Task} & \multirow{2}{*}{Method} & \multirow{2}{*}{Backbone} & \multicolumn{2}{c}{AVSS} \\
                      &                         &                           & $\mathcal {M_J}$           & $\mathcal {M_F}$          \\ \midrule
\multirow{2}{*}{VOS}  & 3DC~\cite{mahadevan2020making}                    & resnet18                  & 17.3       & 21.6       \\
                      & AOT~\cite{yang2021associating}                     & resnet50                  & 25.4        & 31.0         \\ \midrule
\multirow{2}{*}{AVSS} & TPAVI~\cite{zhou2023audio}  & PVT-v2                    & 29.8       & 35.2       \\
                      & CATR & PVT-v2                    &  {\color{red}32.8}           &    {\color{red}38.5}        \\ \bottomrule
\end{tabular}
\caption{Quantitative comparisons on AVSBench-semantic datasets~(AVSS). {\color{red}{Red}} indicates the best performance.}
\label{tab:sematic}
\vspace{-0.5cm}
\end{table}

\section{Experiment}
\subsection{Implementation Details}
\noindent {\bf Datasets.}
We train and validate our model on three datasets: Semi-supervised Single-sound Source Segmentation~(S4), Fully-supervised Multiple-sound Source Segmentation~(M3), and Fully-supervised Audio-Visual Semantic Segmentation (AVSS). S4 and M3 datasets provide binary segmentation maps identifying the pixels of sounding objects, while the AVSS dataset offers semantic segmentation maps as labels. The S4 dataset contains audio samples with a single target object, supplying ground-truth solely for the initial frame during training. Evaluation necessitates predictions for all video frames in the test set. In contrast, both M3 and AVSS datasets contain audio samples with multiple target objects and furnish ground-truth data for all frames throughout the training phase.

\noindent {\bf Training Details.}
We conduct training and evaluation on S4, M3 and AVSS datasets, with the backbone ResNet-50 and Pyramid Vision Transformer~(PVT-v2). The channel size of the spatial-temporal encoding module is set to $C = 256$. We use the VGGish model to extract audio features and use the Adam optimizer with a learning rate of 1e-5 for the fully-supervised MS3 settings, 1e-4 for the semi-supervised S4 and the fully-supervised AVSS settings. The batch size is set to 4 and the number of audio-constrained queries is set as 50. On S4, we set the training epoch as 25, and On M3 and AVSS, we set the training epoch as 100.

\noindent {\bf Evaluation Metrics.}
 We use standard metrics Jaccard index~\cite{everingham2010pascal} $\mathcal J$ and F-score $\mathcal F$ as the evaluation metrics, where $\mathcal J$ and $\mathcal F$ measure the region similarity and contour accuracy, respectively. In our experiment, we use $\mathcal {M_J}$ and $\mathcal {M_F}$ to denote the mean metric values over the whole dataset.

\begin{table}[t]
\label{tab:ablation}
\begin{tabular}{ccccc}
\toprule
\multirow{2}{*}{PVT-v2} & \multicolumn{2}{c}{M3}          & \multicolumn{2}{c}{S4}          \\ 
                        & \multicolumn{1}{c}{$\mathcal {M_J}$}    & $\mathcal {M_F}$    & \multicolumn{1}{c}{$\mathcal {M_J}$}    & $\mathcal {M_F}$    \\ \midrule
Baseline                & \multicolumn{1}{c}{51.2} & 63.3 & \multicolumn{1}{c}{77.9} & 87.6 \\ 
(+) Annotation            & \multicolumn{1}{c}{53.1} & 64.7 & \multicolumn{1}{c}{79.1} & 88.5 \\ 
(+) Decoupled A-V Transformer               & \multicolumn{1}{c}{57.7} & 68.6 & \multicolumn{1}{c}{82.6} & 90.1 \\ 
(+) Blockwise-Encoded Gate                & \multicolumn{1}{c}{60.8} & 70.3 & \multicolumn{1}{c}{83.8} & 91.1 \\
(+) Audio-queried Decoding       & \multicolumn{1}{c}{62.7} & 74.5 & \multicolumn{1}{c}{84.4} & 91.3 \\ \bottomrule
\end{tabular}
\caption{Ablation analysis on the M3 and S4 dataset with PVT-v2 backbone.}
\vspace{-1cm}
\label{tab:ablation}
\end{table}

 \subsection{Comparison}
 \noindent {\bf Compare with Other Tasks.}
Audio-visual video segmentation (AVVS) is a relatively new and emerging task, first introduced by~\cite{DBLP:conf/eccv/ZhouWZSZBGKWZ22}, which aims to segment target objects in videos based on corresponding sounds. Although some well-established tasks, such as sound source localization~(SSL), video object segmentation~(VOS) and salient object detection~(SOD) can perform video object segmentation, we utilize state-of-the-art methods from these related tasks as a comparative benchmark for our experiments. As evident in Table~\ref{tab:all}, there exists a significant performance gap between SSL-based methods and our CATR, primarily due to the lack of pixel-level results in SSL. Furthermore, our model demonstrates a clear advantage over video object segmentation (VOS) and salient object detection (SOD) methods on both S4 and M3 datasets. This superior performance can be attributed to the fact that VOS and SOD are single-mode tasks and do not utilize sound information. In summary, the comparison with SOTA methods from related tasks substantiates the exceptional performance of our model in AVVS.

\noindent {\bf Compare with SOTA TPAVI.}
Our proposed CATR outperforms the previous SOTA TPAVI on all datasets~(S4, M3 and AVSS) with two backbones~(see Figure~\ref{tab:all} and \ref{tab:sematic}). This improvement is due to the integration of the decoupled audio-visual transformer encoding module~(DAVT) and the object-aware audio-queried decoding module. The DAVT block captures the combinatorial dependence of audio and video, combining audio and video in the space dimension to capture the temporal characteristics of this multi-modal combination. Compared to the previous model that considered the audio-visual temporal and interactive features independently, the combinatorial dependence we obtained is better equipped to locate the referred object. Additionally, our object-aware audio-queried decoder utilizes multiple queries containing rich audio cues and object-level information, providing more accurate object segmentation and target location compared to the previous model's decoding directly. By considering both object-level and audio-constrained decoding, our model achieves more precise results.

\noindent {\bf Compare with the Processed Data.}
Limited datasets exist for audio-visual video segmentation, leading ~\cite{DBLP:conf/eccv/ZhouWZSZBGKWZ22} to introduce AVSBench-object datasets first. Among these, S4 represents a semi-supervised learning task, providing ground-truth for the first frame in the training set. To maximize dataset utility without incurring additional labor, we devised a complementary approach for S4 and M3 datasets. Specifically, during M3 training, we employed AOT~\cite{yang2021associating} to predict unlabeled frames of the S4 training set, using these predictions as ground-truth for the AVVS task. Concurrently, we preserved the same setting as TPAVI, implementing semi-supervised training for the S4 dataset and fully supervised training for the M3 dataset.

Table~\ref{tab:all}'s experimental results demonstrate that our model's performance on the original dataset~(CATR) surpasses the previous state-of-the-art TPAVI, and the supplementary labeling method (CATR*) further enhances the model's effectiveness.

\begin{table}[t]
\begin{tabular}{ccccc}
\toprule
\multirow{2}{*}{Resnet50} & \multicolumn{2}{c}{M3} & \multicolumn{2}{c}{S4} \\
                          & $\mathcal {M_J}$           & $\mathcal {M_F}$         & $\mathcal {M_J}$          & $\mathcal {M_F}$         \\ \midrule
TPAVI w audio             & 47.9       & 57.8      & 72.8       & 84.8      \\
CATR w/o audio            & 36.4       & 51.4      & 73.2       & 84.6      \\
CATR w audio              & {\color{red}52.1}       & {\color{red}64.6}      & {\color{red}74.1}       & {\color{red}86.1}      \\ \bottomrule
\multirow{2}{*}{PVT-v2}   & \multicolumn{2}{c}{M3} & \multicolumn{2}{c}{S4} \\
                          & $\mathcal {M_J}$          & $\mathcal {M_F}$         & $\mathcal {M_J}$          & $\mathcal {M_F}$         \\ \midrule
TPAVI w audio             & 54.0         & 64.5      & 78.7       & 87.9      \\
CATR w/o audio            & 43.9       & 57.6      & 80.7       & 89.1      \\
CATR w audio              & {\color{red}62.7}       & {\color{red}74.5}        & {\color{red}84.4}         & {\color{red}91.3}      \\ \bottomrule
\end{tabular}
\caption{Comparison between TPAVI with audio information and CATR without audio information.}
\label{tab:audio}
\vspace{-0.5cm}
\end{table}

\begin{table}[t]
\begin{tabular}{ccc}
\toprule
\multirow{2}{*}{PVT-v2} & \multicolumn{2}{c}{M3} \\
                        & $\mathcal {M_J}$          & $\mathcal {M_F}$         \\ \midrule
CATR                    & {\color{red}62.7}       & {\color{red}74.5}      \\
w/o spatial fusion      & 59.7       & 70.7      \\
w/o temporal A-V fusion & 58.4       & 69.8      \\
w/o temporal V-A fusion & 61.8       & 71.0        \\ \bottomrule
\end{tabular}
\caption{Ablation analysis of spatial-temporal encoding.}
\label{tab:spatial}
\vspace{-0.5cm}
\end{table}

\begin{table}[t]
\begin{tabular}{ccccc}
\hline
\multirow{2}{*}{Gate   Channel} & \multicolumn{2}{c}{M3} & \multicolumn{2}{c}{S4} \\
                                & $\mathcal {M_J}$          & $\mathcal {M_F}$         & $\mathcal {M_J}$          & $\mathcal {M_F}$         \\ \hline
2                               & 58.4       & 69.8      & 83.5       & 90.8      \\
64                              & 61.8       & 70.9      & 83.9       & 91.1      \\
128                             & 62.1       & 72.9      & 84.2       & 91.2      \\
256                             & 62.7       & 74.5      & 84.4       & 91.3      \\ \hline
\end{tabular}
\caption{Analysis of the number of channels in Blockwise-Encoded Gate with PVT-v2 backbone.}
\label{tab:channel}
\vspace{-0.8cm}
\end{table}

\subsection{Contribution of The Core Components}
Table~\ref{tab:ablation} demonstrates the contributions of each proposed module to the overall performance enhancement in CATR, utilizing PVT-v2 and ASPP as encoding and expanded fused feature maps as decoding in the baseline. Given the limited original training samples, it is essential to maximize the use of available data. We augment the two training sets, respecting their semi-supervised and fully supervised configurations, due to the similarity of their segmentation objectives. Specifically, we incorporated M3 video data into the S4 training set and supplemented the S4 training set with AOT-generated ground-truth when training the M3 dataset. The second row in Table~~\ref{tab:ablation} indicates that our additional annotation improves the performance of both M3 and S4 datasets.

Furthermore, our experiments reveal that the decoupled audio-visual transformer encoding, blockwise-encoded gate, and audio-queried decoding modules significantly enhance the model's performance. Notably, the audio-queried decoding module exhibits a more substantial improvement in the M3 dataset than in the S4 dataset~($\mathcal {M_J}$ is up 6.5 vs. 0.6). This is attributable to the multiple objectives in M3 dataset videos, which complicate segmentation target identification. The audio-constrained query contains rich object-level information and guides the segmentation effectively.

\begin{figure*}[t]
    \centering
    \includegraphics[width=0.95\textwidth]{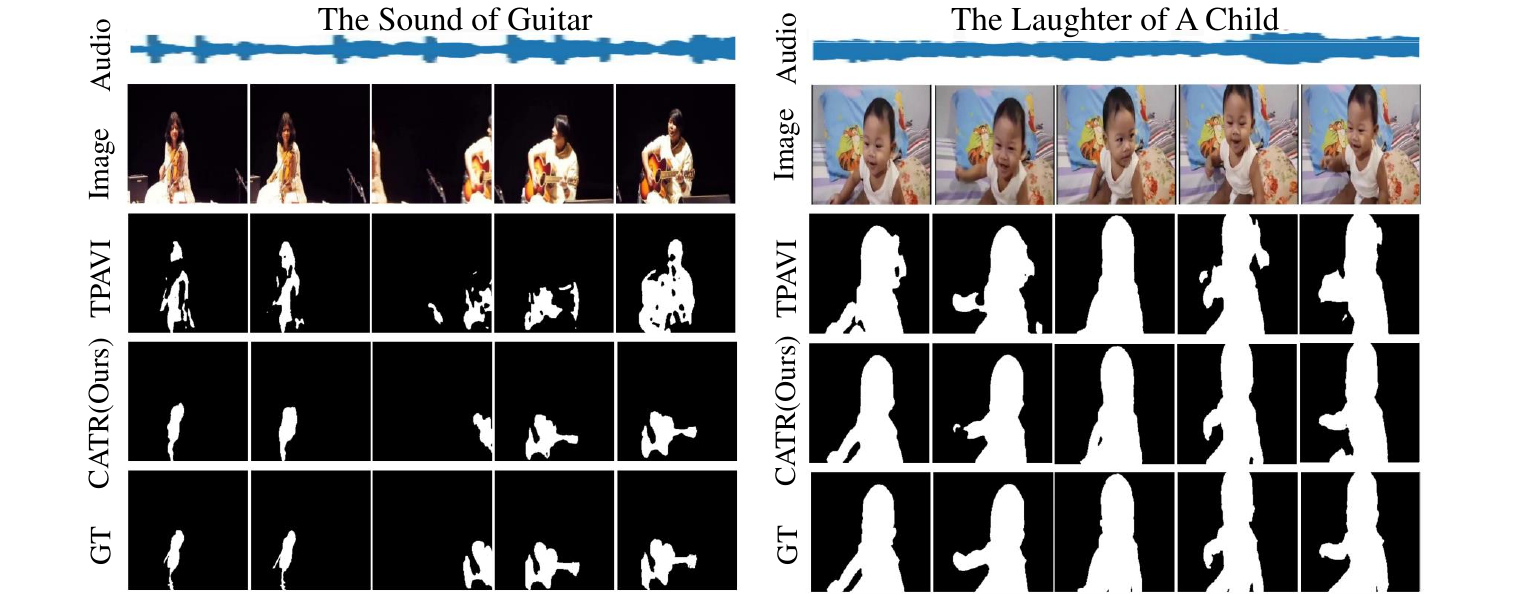}
    \vspace{-0.2cm}
    \caption{{\bf Comparative analysis of the TPAVI method and our proposed CATR. We present two qualitative examples from the M3 and S4 datasets. The M3 dataset example~(left) demonstrates TPAVI's inability to detect the transition of auditory objects, such as from a violin to a guitar, whereas CATR accurately predicts these changes in alignment with the audio signal. In S4 example~(right), CATR exhibits better performance on pixel-level segmentation in the presence of a complex background.} }
    \label{fig:visual-2}    
\end{figure*}

\noindent {\bf The Impact of Decoupled A-V Transformer Encoding.}
We developed a decoupled spatial-temporal encoding block consisting of three components. Initially, we integrated audio and video features in the spatial domain using a spatial fusion method, capturing the temporal dependence of this combination. Subsequently, the spatially fused features were processed through both the temporal A-V and temporal V-A modules. Table~\ref{tab:spatial} demonstrates the contributions of each component to overall performance, highlighting the critical role played by the temporal A-V module. We attribute its significance to the predominant use of video features in the final decoding process, where video features serve as the key and value within the temporal A-V module, preserving crucial video information. To further illustrate this, we examined the attention maps of features processed by the spatial-temporal encoding block, depicted in Figure~\ref{fig:attention}. The initial spatial fusion attention map appears scattered, particularly in the second example, indicating insufficient integration of audio guidance information. In contrast, attention maps for both temporal A-V and temporal V-A modules are more precise and focused, with the temporal A-V map in the second example almost exclusively centered on the piano, underscoring its importance.

\noindent {\bf The Impact of Blockwise-Encoded Gate.}
To optimize the contributions from each encoder block, the Blockwise-Encoded Gate was devised to fully harness the potential of the individual encoders' features. Table~\ref{tab:ablation} demonstrates the enhancement in model performance when incorporating the Blockwise-Encoded Gate. Table~\ref{tab:channel} examines the influence of varying the number of channels within the Blockwise-Encoded Gate, where channels denote the quantity obtained after passing through a convolutional layer, reflecting the designated number of weights. The experimental findings indicate that optimal performance is achieved with 256 channels, corresponding to our feature dimension. This suggests that assigning a weight to each feature channel enables the model to effectively account for the proportional contribution of each feature.

\noindent {\bf The Impact of Audio-Queried Decoding.}
In the decoding phase, we developed $N$ learnable queries incorporating auditory cues and comprehensive object-level information. We employed the $\mathcal{C}_{\text {match }}$ function to select the query optimally aligned with the audio features, which served as the final mask. As demonstrated in Table 3, our audio-constrained query decoding approach substantially enhances the model's performance. Previous models neglect audio information in their decoding stages, resulting in segmentation outcomes predominantly influenced by adjacent video frames. Consequently, by emphasizing audio features during the decoding process, we effectively improve overall performance.

\subsection{The Impact of Audio Signals}
The enhanced performance of CATR prompts an inquiry: Does this improvement stem from a superior comprehension of pixel-level video features or more effective utilization of audio features? To investigate, we conducted an experiment that removed audio features from the Spatial-Temporal Encoding Module and applied self-attention to video features. Additionally, we replaced learnable queries, originally constrained by audio, with video features in the decoding module. Table~\ref{tab:audio} presents the results of CATR without audio. These findings reveal that (1) our model effectively leverages audio features, as evidenced by the significant improvement in the M3 dataset when comparing CATR with and without audio~($\mathcal {M_J}$ is 0.627 vs. 0.439 with PVT-v2); and (2) our model demonstrates a more advanced understanding of pixel-level video features, as shown by the superior performance in the S4 dataset, even without employing audio information, surpassing the previous state-of-the-art TPAVI~($\mathcal {M_J}$ is 0.807 vs. 0.787 with PVT-v2).

\section{Conclusion}
We introduce a novel Combinatorial-Dependence Audio-Queried Transformer~(CATR) framework that achieves state-of-the-art performance on all three datasets using two backbones. Unlike previous methods that treated temporal video information and audio-visual interaction separately, our proposed combinatorial dependence fusion approach comprehensively accounts for the spatial-temporal dependencies of audio-visual combination. Additionally, we propose the audio-constrained learnable queries to incorporate audio information comprehensively during decoding. These queries contain object-level information that can select which object is being referred to segment. To further enhance performance, we introduce a blockwise-encoded gate that balances contributions from multiple encoder blocks. Our experimental results demonstrate the significant impact of these novel components on overall performance. 

\noindent{\bf Limitations:} Objects with similar auditory characteristics can confound video segmentation outcomes when they coexist within a single frame. To address this challenge, we plan to explore the refinement of audio feature pre-processing in future research. {\bf Broader Impact:} The exceptional performance of CATR enables its practical implementation in audio-guided video segmentation applications. These applications include utilizing auditory cues to accentuate objects in augmented and virtual reality environments and generating pixel-level object maps for surveillance inspections. We expect that our research will contribute to practical applications of audio-guided video segmentation.

\begin{acks}
This work was supported by the Fundamental Research Funds for the Central Universities~(No.~226-2023-00048), the National Key Research \& Development Project of China~(2021ZD0110700), and the National Natural Science Foundation of China~(U19B2043, 61976185). 
\end{acks}

\clearpage
{\small
\bibliographystyle{ACM-Reference-Format}
\balance
\bibliography{acmart}
}

\end{document}